\documentclass[final,3p,12pt]{elsarticle}
\usepackage{amssymb}
\usepackage{amsmath}
\usepackage{adjustbox}
\usepackage{mathtools}
\usepackage{xcolor}
\usepackage{soul}
\usepackage{graphicx} % Required for inserting images
\usepackage[linesnumbered,ruled,lined]{algorithm2e}
\usepackage{bm}
\usepackage{bbm}

\newcommand{\Real}{\mathbb{R}}

\newcommand\vb{\mathbf{b}}

\newcommand\vh{\mathbf{h}}

\newcommand\vx{\mathbf{x}}

\newcommand\vtheta{\bm{\theta}}

\newcommand\mc{\mathbf{C}}

\newcommand\mf{\mathbf{F}}
\newcommand\mg{\mathbf{G}}

\newcommand\mi{\mathbf{I}}

\newcommand\mo{\mathbf{O}}

\newcommand\mU{\mathbf{U}}

\newcommand\mw{\mathbf{W}}
\newcommand\mx{\mathbf{X}}

\SetKwInput{KwInput}{Input}                % Set the Input
\SetKwInput{KwOutput}{Output}

\begin{document}

\begin{frontmatter}

\title{Deep Learning-Based Residual Useful Lifetime Prediction for Assets with Uncertain Failure Modes}

\author[inst1]{Yuqi Su}

\affiliation[inst1]{organization={Operations Research Graduate Program, North Carolina State University},%Department and Organization
            addressline={915 Partners Way}, 
            city={Raleigh},
            postcode={27607}, 
            state={NC},
            country={USA}}

\author[inst2]{Xiaolei Fang}

\affiliation[inst2]{organization={Edward P. Fitts Department of Industrial and Systems Engineering, North Carolina State University},%Department and Organization
            addressline={915 Partners Way}, 
            city={Raleigh},
            postcode={27607}, 
            state={NC},
            country={USA}}

\begin{abstract}

Industrial prognostics focuses on utilizing degradation signals to forecast and continually update the residual useful life of complex engineering systems. However, existing prognostic models for systems with multiple failure modes face several challenges in real-world applications, including overlapping degradation signals from multiple components, the presence of unlabeled historical data, and the similarity of signals across different failure modes. To tackle these issues, this research introduces two prognostic models that integrate the mixture (log)-location-scale distribution with deep learning. This integration facilitates the modeling of overlapping degradation signals, eliminates the need for explicit failure mode identification, and utilizes deep learning to capture complex nonlinear relationships between degradation signals and residual useful lifetimes. Numerical studies validate the superior performance of these proposed models compared to existing methods.

\end{abstract}


\end{frontmatter}

\section{Introduction}\label{section1}
Degradation is a gradual and irreversible process of damage accumulation that finally results in the failure of engineering systems/assets. Although underlying degradation processes are usually difficult to observe, there are often some manifestations associated with physical degradation processes that can be monitored using sensor technology. The observed sensing data are known as degradation signals, which can be used to predict the residual useful life (RUL) of assets via prognostic modeling. Prognostic is usually achieved by developing statistical/machining learning methods that map an asset's degradation signals to its time-to-failure (TTF) or RUL. Numerous prognostic models exist in the literature, some tailored for applications with a single failure mode \cite{gebraeel2005residual, fang2017multistream, zhou2023supervised, zhou2023federated,shi2021remaining,su2024two, he2021digital,
ibrahim2024integrated}, while others cater to systems with multiple failure modes \cite{chehade2018data,kundu2019multiple,ragab2019prognostics}.

Existing prognostic models for systems with multiple failure modes typically follow a two-stage approach, starting with the classification or identification of failure modes based on degradation signals, followed by constructing a prognostic model for each identified failure mode to predict the RUL. However, this method encounters several practical challenges in real-world applications. \textit{The first major challenge is the overlapping degradation signals due to ambiguous degradation sources}. Degradation signals often originate from sensors installed on the exteriors of machines, which may capture overlapping degradation effects from multiple internal components. For instance, a vibration sensor on a gearbox might detect signals influenced by both gears and bearings, complicating the task of attributing the degradation to a specific source. Although some feature extraction methods can be applied to isolate signals for each failure mode, few can perfectly decouple the sources of degradation. \textit{The second challenge arises from the lack of labeled training data}. In many industrial settings, historical degradation data often lack labels indicating failure modes, leaving the corresponding failure modes of observed signals unclear. This absence of labels complicates the training of effective classification models, as it is challenging to definitively identify the specific failure mode associated with each signal. \textit{The third challenge stems from the similarity of degradation signals among different components or failure modes}. In some cases, especially in the early stages of degradation, the signals from various failure modes are so similar that distinguishing them based solely on sensor data becomes difficult.

To overcome the challenges outlined above, this article proposes two prognostic methods that are able to model overlapping degradation signals and eliminate the need for identifying specific failure modes. The proposed methods work by mapping an asset's single or multiple stream time series-based degradation signals to its failure time via deep learning structures. To model overlapping degradation signals and bypass the necessity of specifying failure modes, the proposed methods employ a mixture (Log)-Location-Scale (LLS) distribution to model the failure times. LLS distribution is a family of probability distributions parameterized by a location parameter and a non-negative scale parameter, which have been widely used in reliability engineering to model the failure time of engineering assets \cite{meeker2014statistical}. It includes a variety of distributions that cover most of the failure time distributions of engineering assets in real-world applications. Some examples of LLS distributions include {Normal}, {Log-Normal}, {Logistics}, {Log-Logistics}, {Smallest Extreme Value (SEV)}, and {Weibull}. The mixture LLS distribution is a mixture of two or more LLS distributions, which is an ideal distribution to model two or more groups of failure times such as the failure time of an asset with multiple failure modes \cite{taylor1995semi}. Employing the mixture LLS distribution enables the construction of a prognostic model that effectively utilizes overlapping degradation signals and obviates the need for identifying specific failure modes.

The first prognostic model we propose, referred to as Deep Learning-Based Prognostic 1 (DLBP1), operates as follows. Degradation signals are first preprocessed using a Sliding Window Method (SWM) (more details later in Section \ref{section32}) to make sure that the data from all the assets are of the same length, which is required by the subsequent analysis. The degradation signals across various assets often differ since an asset's degradation signals are usually truncated by its failure time (i.e., no degradation data can be observed beyond the failure time since the asset is stopped for maintenance or replacement once it is failed), and the failure time varies from one asset to another. The SWM addresses the varying-length challenge and provides processed degradation signals with the same length. The processed degradation signals then serve as the input of Long Short-Term Memory (LSTM) neural networks  \cite{hochreiter1997long}, which is followed by Fully Connected (FC) layers. The output of the set of FC layers is then connected to another layer, whose neurons represent the parameters of mixture LLS distributions. The parameters are then used to calculate the assets' RUL. For example, if we consider an asset with two failure modes, and whose RUL can be characterized using a mixture log-normal distribution, the distribution parameter layer consists of six neurons that represent the means, variances, and weights of the two log-normal distributions (i.e., $\mu_1$, $\sigma_1^2$, and $\lambda$ for the first distribution and $\mu_2$, $\sigma_2^2$, and $1-\lambda$ for the second distribution). As a result, the asset's failure time distribution can be calculated as follows: $\lambda\mathcal{LN}(\mu_1,\sigma_1^2)+(1-\lambda)\mathcal{LN}(\mu_2,\sigma_2^2)$. The parameters of the proposed prognostic model can be estimated using a historical dataset that comprises the degradation signals and failure times of a certain number of failed assets. After that, the real-time degradation signals of a partially degraded in-field asset are acquired and fed into the trained prognostic model, and the failure time distribution of the asset is estimated. 

The second model we propose, named Deep Learning-Based Prognostic 2 (DLBP2), resembles DLBP1 but with a key modification: it permits all assets to share the same scale parameter. This approach draws inspiration from a common assumption in statistical learning. For instance, in linear regression, it is typically assumed that while the mean of each response variable is influenced by its corresponding predictors, the variance remains consistent across all samples. In other words, DLBP1 allows each asset to have its own unique location and scale parameters, whereas DLBP2 permits each asset to have its own unique location parameter but shares a common scale parameter across all assets. This is achieved by designing a hierarchical parameter estimation algorithm, in which the location and scale parameters are optimized hierarchically. Specifically, only the location parameters are represented by the neurons of the last layer of the deep learning structure. In each iteration, after the weights of the neurons are computed, the scale parameters are optimized separately (more details will be discussed in Section \ref{sectiondl2}). Taking the mixture log-normal distribution as an example again, the distribution parameter layer of DLBP2 is composed of four neurons that represent the means and weights (i.e., $\mu_1$ and $\lambda$ for the first distribution and $\mu_2$ and $1-\lambda$ for the second distribution). The shared attributes, scale parameters ($\sigma_1$ and $\sigma_2$), are optimized individually using the neuron outputs from the parameter layer and the training failure times. The updated scale parameters are then propagated into the ensuing round of iteration until the convergence condition is satisfied. 

The proposed prognostic methods integrate mixture LLS distributions with deep learning. We choose deep learning to map degradation signals to failure times due to its superior performance in capturing complex nonlinear relationships between input and output variables, specifically between degradation signals and failure times in prognostics \cite{da2020remaining,nguyen2022probabilistic, nejjar2024domain, zhang2023variational}. The literature features numerous deep learning-based prognostic models suitable for applications with multiple failure modes, some of which do not require the identification of specific failure modes \cite{malhotra2016multi,kim2020bayesian,ellefsen2019remaining}. Most of these models utilize Convolutional Neural Networks (CNNs), Recurrent Neural Networks (RNNs), or LSTM, a special variant of RNN proposed by \cite{hochreiter1997long}. For example, the authors in \cite{yuan2016fault} constructed an LSTM deep neural network for RUL prediction and showed its enhanced performance over other RNN networks such as GRU-LSTM and AdaBoost-LSTM. Article \cite{malhotra2016multi} proposed an LSTM Encoder Decoder-based model to estimate the health index of engineering assets and validated its performance in capturing the severity of faults. Research paper \cite{da2020remaining} proposed LSTM-DANN network for RUL prediction. In article \cite{kim2020bayesian}, the authors developed a Bayesian deep learning framework to predict RULs. Also, in article \cite{liu2022aircraft}, a double attention-based framework was proposed for aircraft engine RUL prognostics. \textit{However, few existing models have specifically designed a deep learning structure to address the overlapping effects in degradation signals}. The methods proposed in this article integrate mixture LLS distributions into deep learning, which enables the modeling of overlapping degradation signals since mixture LLS distributions intrinsically are capable of modeling signals coupled from multiple degradation processes. In addition, LLS distributions represent valuable domain knowledge since they cover most of the failure time distributions of engineering assets in real-world applications. By integrating LLS distribution into deep learning, we expect that the prediction performance of our proposed prognostic models can be enhanced. This enhancement will be verified through numerical studies, as detailed in Section \ref{section4} later on.

The rest of the paper is organized as follows. Section \ref{section3} discusses the details of the proposed prognostic methods. Next, Section \ref{section4} evaluates the performance of the proposed models using aircraft turbofan engine degradation data from a physics-based simulation model developed by NASA. Finally, Section \ref{section5} concludes.

\section{The Deep Learning-Based Prognostic Method}\label{section3}

In this section, we discuss the details of our deep learning-based prognostic methods for the RUL prediction of complex engineering assets with multiple failure modes. We first introduce the mixture LLS distribution that is used to characterize the failure time distribution. Then, we present how to integrate mixture LLS distributions with deep learning.

\subsection{Mixture LLS Distribution}\label{section31}

The LLS distribution is a family of probability distributions parameterized by a location parameter and a non-negative scale parameter. It includes a variety of distributions that are widely used in reliability engineering and survival analysis. For a random variable $Y$ from a location-scale distribution, its probability density function (i.e., pdf) can be expressed as $f(Y;\mu,\sigma)=\frac{1}{\sigma}h(\frac{Y-\mu}{\sigma})$, where $\mu$ is the location parameter, $\sigma>0$ is the scale parameter, and $h(\cdot)$ is a standard pdf of the location-scale distribution. For example, $h(\epsilon)=\exp(\epsilon-\exp(\epsilon))$ for SEV
distribution, $h(\epsilon)=\exp(\epsilon)/(1+\exp(\epsilon))^{2}$
for logistic distribution,
and $h(\epsilon)=1/\sqrt{2\pi}\exp(-\epsilon^{2}/2)$
for normal distribution.
A random variable ${Y}$ is a member of the LLS family if $\log({Y})$ belongs to the location-scale family. The pdf of a random variable $Y$ from the LLS family can be denoted as $f(Y;\mu,\sigma)=\frac{1}{Y\sigma}h(\frac{\log(Y)-\mu}{\sigma})$.

A mixture LLS distribution is a weighted combination of two or more specific distributions that come from the LLS family. Considering an asset with $K$ failure modes, the pdf of its failure time can be denoted as follows:

\begin{equation}\label{equation311}
    g(y;\mu_1,\cdots,\mu_K,\sigma_1,\cdots,\sigma_K)=\sum_{k=1}^K\lambda_kf_k(y;\mu_k,\sigma_k)
    ,
\end{equation}
where $y$ represents the failure time of the asset, $f_k(\cdot)$ is the pdf of the $k$th probability distribution, $\mu_k$ and $\sigma_k$ are the location and scale parameter of the $k$th distribution, respectively. $\lambda_k$ is the weight coefficient, and $\sum_{k=1}^K\lambda_k=1$. Below we give some examples of the mixture LLS distribution:

\subsubsection{Mixture Log-Normal Distribution}
\begin{equation}\label{equation312}
\begin{aligned}
    &g(y;\mu_1,\cdots,\mu_K,\sigma_1,\cdots, \sigma_K)=&\sum_{k=1}^K\lambda_k\frac{1}{y\sigma_k\sqrt{2\pi}}\exp{\{-\frac{(\log y-\mu_k)^2}{2\sigma_k^2}\}},
    \end{aligned}
\end{equation}
where $\mu_k\in\mathbb{R}, \sigma_k>0\  \text{for}\ k=1,2,\cdots, K$, and the mean is $\sum_{k=1}^K\lambda_k\exp{\{\mu_k+\frac{\sigma^2_k}{2}\}}$. \\

\subsubsection{Mixture Weibull Distribution}
\begin{equation}\label{equation313}
 \begin{aligned}
    &g(y;\mu_1,\cdots,\mu_K,\sigma_1,\cdots, \sigma_K)=&\sum_{k=1}^K\lambda_k\frac{\mu_k}{\sigma_k}(\frac{y}{\sigma_k})^{\mu_k-1}\exp{\{-(\frac{y}{\sigma_k})^{\mu_k}\}}
    ,
    \end{aligned}
\end{equation}
where $\mu_k>0, \sigma_k>0$ for $k=1,2,\cdots, K$, and the mean is $\sum_{k=1}^K\lambda_k\sigma_k\Gamma(1+\frac{1}{\mu_k})$. \\

\subsubsection{Mixture Log-Logistic Distribution}
\begin{equation}\label{equation314}
  \begin{aligned}
    &g(y;\mu_1,\cdots,\mu_K,\sigma_1,\cdots, \sigma_K)=&\sum_{k=1}^K\lambda_k\frac{\sigma_k}{\mu_k}(\frac{y}{\mu_k})^{\sigma_k-1}\{(1+(\frac{y}{\mu_k})^{\sigma_k})^2\}^{-1}
    ,
    \end{aligned}
\end{equation}

\noindent where $\mu_k>0$ when $\sigma_k>1$ for $k=1,2,\cdots, K$ and undefined otherwise, and the mean is $\sum_{k=1}^K\lambda_k\frac{\mu_k\pi}{\sigma_k}\{\sin{(\frac{\pi}{\sigma_k})}\}^{-1}$.

\subsubsection{Mixture Log-Normal and Weibull Distribution}
\begin{equation}\label{equation314plus}
   \begin{aligned}    &g(y;\mu_{1,1},\cdots,\mu_{1,K_1},\mu_{2,1},\cdots,\mu_{2,K_2},\sigma_{1,1},\cdots,\sigma_{1,K_1},\sigma_{2,1},\cdots, \sigma_{2,K_2})
    \\=&\sum_{k_1=1}^{K_1}\lambda_{k_1}\frac{1}{y\sigma_{1,k_1}\sqrt{2\pi}}\exp{\{-\frac{(\log y-\mu_{1,k_1})^2}{2\sigma_{1,k_1}^2}\}}+
    \\&\sum_{k_2=1}^{K_2}\lambda_{K_1+k_2}\frac{\mu_{2,k_2}}{\sigma_{2,k_2}}(\frac{y}{\sigma_{2,k_2}})^{\mu_{2,k_2}-1}\exp{\{-(\frac{y}{\sigma_{2,k_2}})^{\mu_{2,k_2}}\}}
    ,
    \end{aligned}
\end{equation}

\noindent where $\mu_{1,k_1}\in\mathbb{R}, \sigma_{1,k_1}>0\  \text{for}\ k_1=1,2,\cdots, K_1$, and $K_1$ is the number of log-normal distributions. $\mu_{2,k_2}>0,\sigma_{2,k_2}>0$ for $k_2=1,2,\cdots, K_2$. $K_2$ is the number of Weibull distributions. The expectation of the mixture distribution is $\sum_{k_1=1}^{K_1}\lambda_{k_1}\exp{\{\mu_{1,k_1}+\frac{\sigma^2_{1,k_1}}{2}\}}+\sum_{k_2=1}^{K_2}\lambda_{K_1+k_2}\sigma_{2,k_2}\Gamma(1+\frac{1}{\mu_{2,k_2}})$.

\subsection{Deep Learning Model Construction}\label{section32}

The first proposed prognostic model maps the degradation signals of an asset to its RUL using a deep learning structure, which consists of LSTM layers, FC layers, and a distribution parameters layer. Since the lengths of various assets' degradation signals may differ, we first apply an SWM on the original signals such that the lengths of the processed signals from different assets are tailored to be the same. Then, the processed same-length signals are fed into the deep learning framework, the outputs of which are the parameters (i.e., weights, locations, and scales) of the mixture LLS distribution. Finally, the parameters of the mixture distribution are used to calculate the RUL of assets. The second prognostic model is similar to the first one. The main difference between the two models is that the first model allows each asset to have its own location and scale parameters, while the second method only allows each asset to have its own location parameter but requires all assets to share the same set of scale parameters. 
 
\subsubsection{SWM}\label{section321}
    \begin{figure*}[!t]
    \centering
	\includegraphics[width=3.7in]{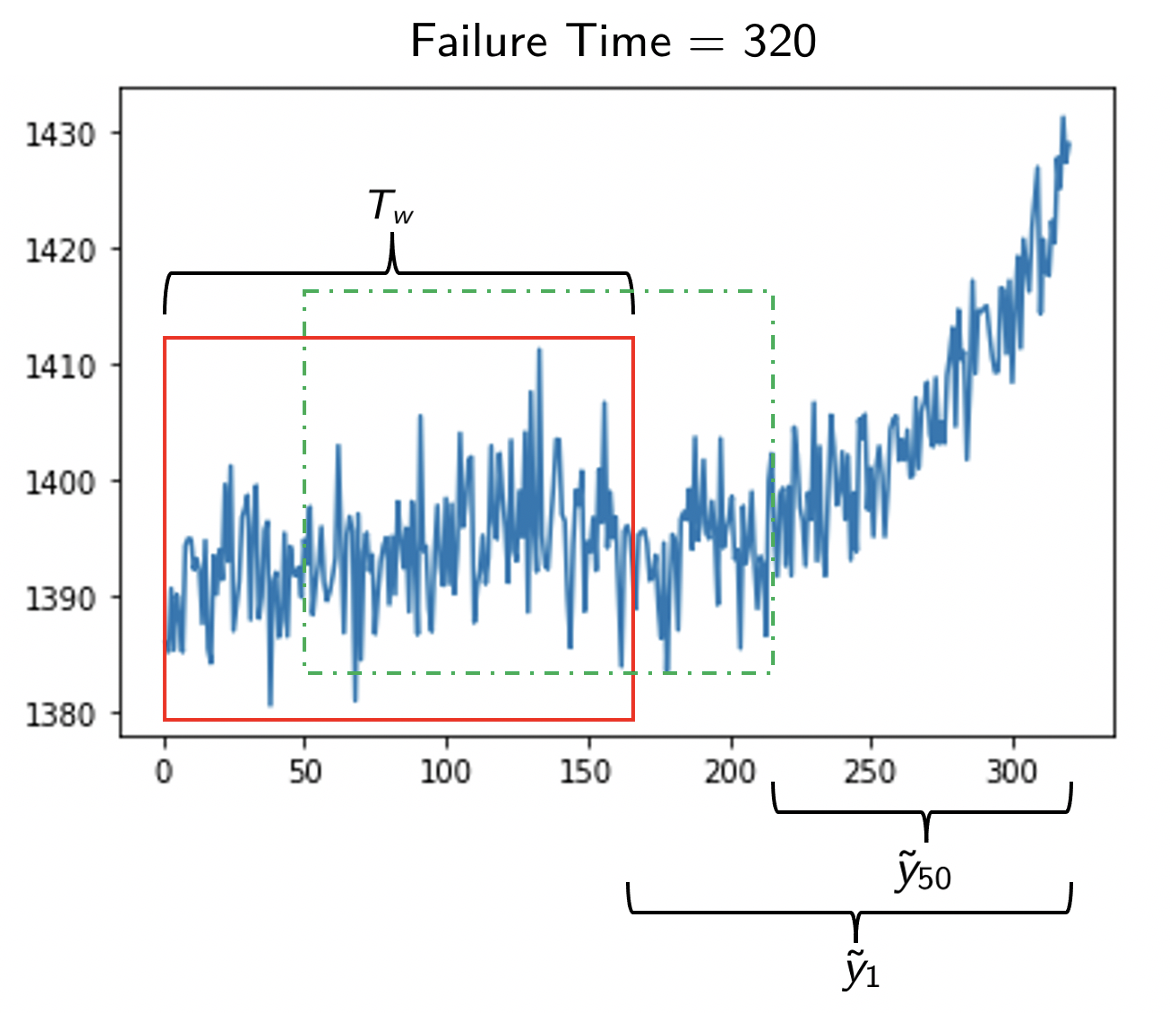}
	  \caption{A demonstration of the SWM. The time-series data is from sensor $\#4$ of engine $\#27$ in the FD003 training dataset, whose real RUL is $320$. The sliding window width is $T_w$. The solid rectangle in red and the dashed rectangle in green represent two truncated signals, and their corresponding RULs are $\tilde{y}_1$ and $\tilde{y}_{50}$, respectively (more details about the dataset will be provided in Section 3). }
	\label{sliding}
     \end{figure*}
     
We assume there exists a historical dataset for model training. The training dataset consists of degradation signals and the failure time of $N$ assets. We denote the RUL and degradation signals of asset $i$ as $y_i$ and $\mathbf{X}_i$, respectively. Here $y_i\in\Real_+$, $\mathbf{X}_i\in\Real^{n_i\times P}$, $n_i$ is the length of degradation signals of asset $i$, $P$ is the number of sensors, and $i=1,\ldots, N$. To ensure the degradation signals from different assets are of the same length, we use a window with width $T_w$ to truncate the signals, as shown in Figure \ref{sliding}. Specifically, we generate the first truncated signal matrix $\boldsymbol{\tilde{X}}_i^{(1)}\in\Real^{T_w\times P}$ by keeping the first $T_w$ rows of the degradation signal matrix $\mathbf{X}_i$, i.e., $\boldsymbol{\tilde{X}}_i^{(1)}=\mathbf{X}_i[1:T_w,1:P]$. The first RUL corresponding to $\boldsymbol{\tilde{X}}_i^{(1)}$ is $\tilde{y}_i^{(1)}=y_i-T_w$. Then, the second truncated signal matrix is generated by keeping the second to the $(T_w+1)$th row of $\mathbf{X}_i$, i.e., $\boldsymbol{\tilde{X}}_i^{(2)}=\mathbf{X}_i[2:T_w+1,1:P]$, and its corresponding RUL is $\tilde{y}_i^{(2)}=y_i-T_w-1$. This operation is repeated until the last truncated signal matrix and its corresponding positive RUL are generated. The SWM is applied to degradation signals of all $N$ assets in the training dataset and the resulting truncated degradation signal matrices and RULs are respectively denoted by $\tilde{\mathbf{X}}_i$ and $\tilde{y}_i$, $i=1,\ldots, n$, where $n$ is the number of samples (which we still call ``assets") after truncation. A typical choice for the width of the sliding window $T_w$ is the smallest signal length among the $N$ assets in the training data set, i.e., $T_w=\inf(\{n_i\}_{i=1}^{N})$. The length may also be selected using model selection methods such as cross-validation. If there exists an asset $i$ whose signal length is smaller than the window width, i.e., $n_i<T_w$, we may apply left zero padding \cite{da2020remaining} to keep the sample size consistent for all the truncated signal matrices.

\subsubsection{LSTM}

To map the truncated degradation signal matrices to the RULs, one viable and widely adopted method is RNN \cite{rumelhart1986learning}, a technique extensively employed for processing sequential data by introducing recurrence that connects the hidden layer with the feedback of the output.
However, vanilla RNN does not always perform well. This is because when the network becomes complex or the time steps between relevant information become longer, the performance of RNN degrades due to the vanishing or exploding gradients issues \cite{hochreiter2001gradient}. To deal with the long-term dependency challenge, the LSTM method \cite{hochreiter1997long}, which is a special type of RNN, is proposed. There are various LSTM variants \cite{greff2016lstm}, and we will use the basic LSTM network \cite{hochreiter1997long} in this article. 

\begin{figure*}[!t]
    \centering
	\includegraphics[width=6in]{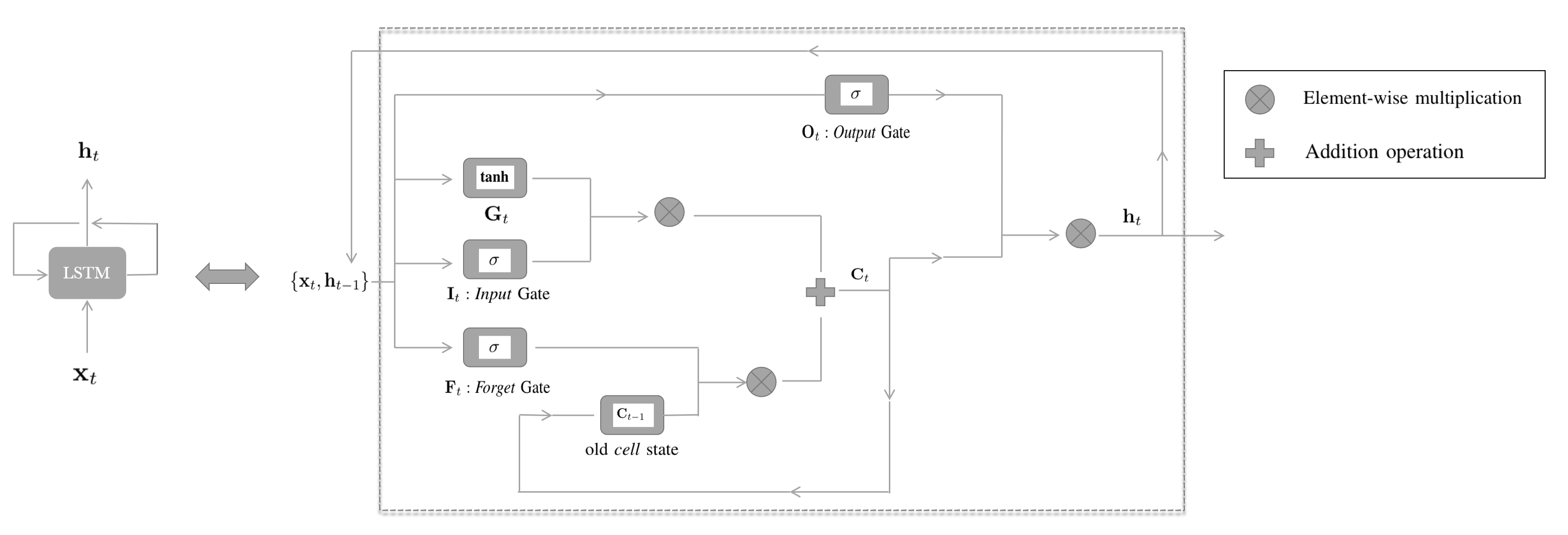}
	  \caption{The LSTM recurrent cell diagram, where $\sigma$ means applying \textit{sigmoid} activation function to the output of the \textit{input} gate, the \textit{forget} gate, as well as the \textit{output} gate, and \textit{tanh} represents the activation function applied to the output of $\mg_t$.}
	\label{lstmflow}
     \end{figure*}

LSTM has the same structure as RNN except that it replaces the hidden layer of RNN with a memory cell, which consists of an \textit{input} gate, a \textit{forget} gate, an \textit{output} gate, and a \textit{cell} state. Figure \ref{lstmflow} shows the detailed recurrent LSTM structure. Let $\vx_t^\top\in\Real^P$ be the rows of $\tilde{\mathbf{X}}_i$, where $t=1,\ldots, T_w$, the LSTM network can be formed as follows \cite{olah2015understanding}:
\begin{equation}\label{equation321}
\begin{aligned}
\mf_t&=\sigma(\mw_f\cdot \vx_t+\mU_f\cdot \vh_{t-1}+\vb_f)\\
\mi_t&=\sigma(\mw_i\cdot \vx_t+\mU_i\cdot \vh_{t-1}+\vb_i)\\
\mg_t&=\tanh(\mw_g\cdot \vx_t+\mU_g\cdot \vh_{t-1}+\vb_g)\\
\mc_t&=\mf_t\otimes \mc_{t-1}+\mi_t\otimes \mg_t\\
\mo_t&=\tanh(\mw_o\cdot \vx_t+\mU_o\cdot \vh_{t-1}+\vb_o)\\
\vh_t&=\mo_t\otimes \tanh(\mc_t),
\end{aligned}
\end{equation}

\noindent where $\vh_{t-1},\vh_t\in\Real^b$ are the outputs of the cell, $b$ is the output dimension of the LSTM network, $\mw_f, \mw_i,$ $\mw_g,\mw_o\in\Real^{m\times P}$ are the weights with respect to the input $\vx_{t}$, $\mU_f,\mU_u,\mU_g,\mU_o\in\Real^{m\times b}$ denote the weight matrices with regard to the output $\vh_{t-1}$ coming from the previous time step, $\vb_f, \vb_i,\vb_g,\vb_o$ are the bias vectors, and $\otimes$ is the element-wise multiplication operator.
$\mc_t$ represents the \textit{cell} state that stores the information added or removed by gate operations. The \textit{forget} gate at time $t$, denoted by $\mf_t$, decides the information to be kept or dropped from the current cell. $\mi_t$ characterizes the \textit{input} gate that determines which information gets updated, and $\mg_t$ uses a nonlinear activation function to regulate and stabilize the output. Denoted by $\mo_t$, the \textit{output} gate determines which information gathered from the \textit{cell} state is going to pass to the next state. Such an LSTM architecture preserves the capability of learning long-term dependencies in sequential data. We refer the readers to the article \cite{hochreiter1997long} for a more detailed introduction to the LSTM network.

\subsubsection{Deep Learning Network Structure Without Shared Scale Parameters}\label{sectiondl1}

In this subsection, we delve into the details of the first DLBP1 model. The degradation signals of an asset are first processed using the previously discussed SWM such that the transformed signals are of the same length. Then, the processed signals are fed into the LSTM layer, which is followed by FC layers. The last FC layer is then connected to a distribution parameters layer, and the neurons of which represent the parameters of the mixture LLS distribution and are used to calculate the RUL of the asset. 

The mapping function and parameters (i.e., weights) of the LSTM layer are denoted by $l(\cdot)$ and $\vtheta_l$, respectively. Similarly,  the mapping function and parameters of the FC layer are respectively represented by $f(\cdot)$ and $\vtheta_f$.  If we consider an asset with $K$ failure modes, there are $3K$ neurons in the distribution parameter layer. The outputs of the first $K$ neurons represent the location parameters of the $K$ LLS distributions, while the outputs of the $K+1,K+2,\ldots, 2K$ neurons stand for the scale parameters. The outputs of the last $K$ (i.e., $2K+1,2K+2,\ldots, 3K$) neurons are the weights corresponding to these $K$ distribution components.

We denote the activation function of the $q$th neuron in the distribution parameter layer as $d_q (\cdot)$ and the neuron weights as $\vtheta_d^q$ (the input of the $q$th neuron is a weighted combination of the outputs of the neurons in the last FC layer), where $q=1,\ldots,3K$. As a result, the output of the $q$th neuron in the distribution parameter layer is $d_q(f(l(\tilde{\mathbf{X}}_i;\vtheta_l);\vtheta_f);\vtheta_d^q)$, where $\tilde{\mathbf{X}}_i$ is the degradation signals of asset $i$, $l(\tilde{\mathbf{X}}_i;\vtheta_l)$ is the output of LSTM layers, $f(l(\tilde{\mathbf{X}}_i;\vtheta_l);\vtheta_f)$ is the output of FC layers. For simplicity, we denote the location parameters corresponding to the first $K$ neurons of the distribution parameter layer as $\mu_{i,k}$'s, where $\mu_{i,k}=d_k(f(l(\tilde{\mathbf{X}}_i;\vtheta_l);\vtheta_f);\vtheta_d^k)$ and $k=1,\ldots,K$. The scale parameters corresponding to the $K+1,K+2,\ldots,2K$ neurons of the distribution parameter layer are denoted by $\sigma_{i,k}$'s, where $\sigma_{i,k}=d_{K+k}(f(l(\tilde{\mathbf{X}}_i;\vtheta_l);\vtheta_f);\vtheta_d^{K+k})$ and $k=1,\ldots,K$. The weights of the $K$ distributions, which are corresponding to the output of the $2K+1,2K+2,\ldots,3K$ neurons in the distribution parameter layer, are denoted as $\tilde\lambda_{i,k}$'s, where $\tilde\lambda_{i,k}=d_{2K+k}(f(l(\tilde{\mathbf{X}}_i;\vtheta_l);\vtheta_f);\vtheta_d^{2K+k})$ and $k=1,\ldots,K$.

The output of the neurons in the distribution parameter layer (i.e., $\mu_{i,k}$, $\sigma_{i,k}$, and $\tilde\lambda_{i,k}$, $k=1,2,\ldots,K$) can be used to calculate the output of the proposed DLBP1 model, which include the weights, location parameters, and scale parameters that can be used to compute the RUL of the $i$th sample. Before doing the calculation, the weights $\tilde\lambda_k$'s need to be normalized such that the summation of the weights from all the distributions equals one. If the normalized weights are denoted by $\lambda_k$'s, the normalization can be accomplished by using $\lambda_{i,k}=\tilde\lambda_{i,k}/\sum_{j=1}^{K}\tilde\lambda_{i,j}$, $k=1,\ldots,K$. As the training dataset $\{\tilde{\mx}_i,\tilde{y}_i\}_{i=1}^n$ being fed into the deep network, the parameters $\{\hat{\mu}_{i,k}, \hat{\sigma}_{i,k}, \hat{\lambda}_{i,k}\}$ for $i=1,\cdots, n$ and $k=1,\cdots, K$ can be estimated.
We use the negative log-likelihood function of the mixture LLS distribution as the loss (objective) function,
which can be minimized using Adam optimizer \cite{kingma2014adam}.

\subsubsection{Deep Learning Network Structure with Shared Scale Parameters}\label{sectiondl2} 

In this section, we provide an in-depth illustration of the second network (DLBP2). Unlike the first model where each asset has its own scale parameter, this model requires that all assets share the same set of scale parameters. Similar to Section \ref{sectiondl1}, if we consider an asset with $K$ failure modes, there are $2K$ neurons in the distribution parameter layer. The outputs of the first $K$ neurons represent the location parameters of the $K$ LLS distributions, and the outputs of the last $K$ (i.e., $K+1,K+2,\ldots, 2K$) neurons are the weights corresponding to these $K$ distribution components. In the training process, following the notation in Section \ref{sectiondl1}, we denote the activation function of the $q$th neuron in the distribution parameter layer as $d_{q}(\cdot)$ and the weights as $\vtheta_{d}^{q}$ (the input of the $q$th neuron is a weighted combination of the outputs of the neurons in the last FC layer), where $q=1,\cdots, 2K$. Then the output of the $q$th neuron in the distribution parameter layer is $d_{q}(f(l(\tilde{\mathbf{X}};\vtheta_{l});\vtheta_{f});\vtheta_{d}^q,\sigma)$. 
As a result, the location parameters for asset $i$ is $\mu_{i,k}=d_{k}(f(l(\tilde{\mathbf{X}}_i;\vtheta_{l});\vtheta_{f});\vtheta_{d}^k,\sigma_{1:K})$ for $k=1,\ldots,K$
and the weights of the $K$ distribution components are $\tilde{\lambda}_{i,k}=d_{K+k}(f(l(\tilde{\mathbf{X}}_i;\vtheta_{l});\vtheta_{f});\vtheta_{d}^{K+k},\sigma_{1:K})$ for $k=1,\ldots,K$ and the normalized distribution weights are denoted as $\lambda_{i,k}$'s, where $\lambda_{i,k}=\tilde\lambda_{i,k}/\sum_{j=1}^{K}\tilde\lambda_{i,j}$, $k=1,\ldots,K$.

By feeding the training dataset $\{\tilde{\mx}_i,\tilde{y}_i\}_{i=1}^n$ and the scale parameters into the deep neuron network, we can estimate the location and weights parameters {$\{\hat{\mu}_{i,1}, \hat{\mu}_{i,2},\cdots, \hat{\mu}_{i,K}\}_{i=1}^n, \{\hat{\lambda}_{i,1}$, $\hat{\lambda}_{i,2}, \cdots, \hat{\lambda}_{i,K}\}_{i=1}^n$.}
As a result, the RUL $\hat{\tilde{y}}_{i}$ can be estimated by using the mean of the mixture distribution. 
Here we give some examples of the estimated RUL $\hat{\tilde{y}}_{i}$ under different mixture distributions. 
\begin{itemize}
    \item  \noindent Mixture log-normal: \begin{equation}\label{al2equ1}
 \hat{\tilde{y}}_{i}=\sum_{k=1}^K\hat{\lambda}_{i,k}\exp\Bigl\{\hat{\mu}_{i,k}+\frac{\hat{\sigma}^2_{k}}{2}\Bigr\}
\end{equation}

\item \noindent Mixture Weibull: \begin{equation}\label{al2equ2}
\hat{\tilde{y}}_{i}=\sum_{k=1}^K\hat{\lambda}_{i,k}\hat{\sigma}^2_{k}\Gamma\left(1+\frac{1}{\hat{\mu}_{i,k}}\right)
\end{equation}

\item \noindent Mixture log-logistic: \begin{equation}\label{al2equ3}
\hat{\tilde{y}}_{i}=\sum_{k=1}^K\hat{\lambda}_{i,k}\frac{\hat{\mu}_{i,k}\pi}{\hat{\sigma}_{k}}\Bigl\{\sin\left(\frac{\pi}{\hat{\sigma}_{k}}\right) 
\Bigr\}^{-1}
\end{equation}

\item \noindent Mixture log-normal and Weibull:
\begin{equation}\label{al2equ4}
\hat{\tilde{y}}_{i}=\sum_{k_1=1}^{K_1}\hat{\lambda}_{i,k_1}\exp\Bigl\{\hat{\mu}_{1,i,k_1}+\frac{\hat{\sigma}^2_{1,k_1}}{2}\Bigr\} + 
\sum_{k_2=1}^{K_2}\hat{\lambda}_{i,K_1+k_2}\hat{\sigma}^2_{1,k_2}\Gamma\left(1+\frac{1}{\hat{\mu}_{1,i,k_2}}\right),
\end{equation}
\end{itemize}

\noindent where $K_1,K_2\in\mathbb{R}^+$ and $K_1+K_2=K$. With the estimated location parameters and distribution weights, we then update scale parameters using maximum likelihood estimation (MLE). For components that follow a log-normal distribution, the scale parameters are updated using the formula below:

\begin{equation}\label{eq:sigma_normal}
    \hat{\sigma}_{k} = \sqrt{\frac{\sum_{i=1}^n(\log \tilde{y}_i-\hat{\mu}_{i,k})^2}{n}}
\end{equation}

\noindent
For Weibull and log-logistic distribution, the scale parameters are updated by solving functions $h(\sigma_k)=0$ and $s(\sigma_k)=0$, respectively, where $h(\sigma_k)$ and $s(\sigma_k)$ are defined below:
\begin{equation}\label{eq:sigma_weibull}
        h(\sigma_{k}) \coloneqq \sum_{i=1}^n\frac{\hat{\mu}_{i,k}(\sigma_{k}^{\hat{\mu}_{i,k}}-\tilde{y_i}^{\hat{\mu}_{i,k}})}{\sigma_{k}^{\hat{\mu}_{i,k}+1}}
\end{equation}

\begin{equation}\label{eq:loglogistic}
    s(\sigma_{k}) \coloneqq \sum_{i=1}^n\Biggl\{
    \log \left(\frac{\tilde{y}_i}{\hat{\mu}_{i,k}}\right)\Biggl[1-\frac{2\left(\frac{\tilde{y}_i}{\hat{\mu}_{i,k}}\right)^{\sigma_{k}}}{1+\left(\frac{\tilde{y}_i}{\hat{\mu}_{i,k}}\right)^{\sigma_{k}}}\Biggr]
    \Biggr\}
    +\frac{n}{\sigma_k}
\end{equation}

\noindent
Since $h(\sigma_k)=0$ and $s(\sigma_k)=0$ do not have a closed-form solution, a numerical approach such as the Newton-Raphson method can be applied. The complete algorithm for the training of DLBP2 is summarized in Algorithm \ref{algo:dlbp2nf}. %In the testing process with test dataset  $\{\tilde{\mx}_i,{y}_i\}_{i=1}^{n_t}$, where $n_t$ is the sample size after SWM, we can compute the estimated RUL $\{\hat{y}_i\}_{i=1}^{n_t}$ using the optimal parameters acquired in the training procedure. 

\begin{algorithm}[!t]
  \DontPrintSemicolon
  \SetAlgoLined
\KwInput{
$\{\tilde{\mx}_i,\tilde{y}_i\}_{i=1}^{n}$: \textit{Training data after applying the SWM}\\
$\hspace{1.6cm}I$: \textit{Maximum iteration number}\\
$\hspace{1.6cm}K:$ \textit{Number of mixture distribution components}\\
$\hspace{1.6cm}\epsilon:$ \textit{Convergence tolerance}
}
\KwOutput{
\textit{The converged  neuron weights \mbox{$\vtheta_{l},\vtheta_{f},\{\vtheta_{d}^q\}_{q=1}^{2K}$} and scale parameters $\sigma_{k}$, $k=1,\ldots,K$}
}
 \caption{DLBP2}%%
\label{algo:dlbp2nf}
\SetKwFunction{FName}{Network Flow}
\SetKwFunction{Equations}{Equations}
  \SetKwProg{Fn}{Function}{:}{}
  \SetKwProg{For}{For}{:}{end}
  \SetKw{Break}{break}
  \SetKwProg{TrainP}{(Training Process)}{:}{}
  \SetKwProg{TestP}{(Testing Process)}{:}{}

Initialize scale parameters $\hat{\sigma}_{1}^{(0)},\hat{\sigma}_{2}^{(0)},\cdots,\hat{\sigma}_{K}^{(0)}\in$ Uniform$(0,1)$.\;

 \For{$j=1\ $\text{to} $I$}{
 Estimate the neuron weights
  $\vtheta_{l}^{(j)},\vtheta_{f}^{(j)},\{\vtheta_{d}^{q(j)}\}_{q=1}^{2K}$.\;

  Estimate the location parameters $\hat{\mu}_{i,k}^{(j)}=d_{k}(f(l(\tilde{\mathbf{X}}_i;\vtheta_{l}^{(j)});\vtheta_{f}^{(j)});\vtheta_{d}^{k(j)},\hat{\sigma}_{1:K}^{(j-1)})$ for $k=1,\ldots,K$.\;
  
  Compute the weights of the $K$ distribution components $\tilde{\lambda}_{i,k}^{(j)}=d_{K+k}(f(l(\tilde{\mathbf{X}}_i;\vtheta_{l}^{(j)});\vtheta_{f}^{(j)});\vtheta_{d}^{K+k(j)},\hat{\sigma}_{1:K}^{(j-1)})$ and normalize them, i.e., $\lambda_{i,k}^{(j)}$'s, where $\lambda_{i,k}^{(j)}=\tilde\lambda_{i,k}^{(j)}/\sum_{j=1}^{K}\tilde\lambda_{i,j}^{(j)}$, $k=1,\ldots,K$. \;

  Update the scale parameters  $\hat{\sigma}_{k}^{(j)}$ using \Equations \eqref{eq:sigma_normal}, \eqref{eq:sigma_weibull}, or \eqref{eq:loglogistic}, $k=1,\ldots,K$.\;

  \If{$\frac{1}{K}\sum_{k=1}^K\|\hat{\sigma}_{k}^{(j)}-\hat{\sigma}_{k}^{(j-1)}\|^2<\epsilon$}{\Break}
  }

  Assign $\{\hat{\mu}_{i,k}\}_{i=1}^n = \{\hat{\mu}_{i,k}^{(j)}\}_{i=1}^n, \{\hat{\lambda}_{i,k}\}_{i=1}^n = \{\hat{\lambda}_{i,k}^{(j)}\}_{i=1}^n$ for $k=1,\cdots, K$; and $\vtheta_{l} = \vtheta_{l}^{(j)}, \vtheta_{f} = \vtheta_{f}^{(j)}, \{\vtheta_{d}^{q}\}_{q=1}^{2K} = \{\vtheta_{d}^{q(j)}\}_{q=1}^{2K}.$
  
\end{algorithm}

\subsubsection{Real-Time RUL Prediction}
\color{black}
Once the parameters of the DLBP models are estimated, the model can be used to predict the RUL of a partially degraded in-field asset in real time. Specifically, the in-field asset's real-time degradation signals are first acquired using condition monitoring sensors. Next, the degradation signals are processed using the SWM discussed in Section \ref{section321}, such that the latest 
%$T_w$ 
observations are kept as the truncated signals. Then, the truncated signals are fed into the trained deep learning model, which provides the estimation of the parameters of the mixture LLS distribution corresponding to the in-field asset. Since the weights, location, and scale parameters of the mixture distribution are known, the estimated RUL of the asset can also be estimated.

\section{Numerical Study}\label{section4}

In this section, we evaluate the performance of the proposed deep learning-based prognostic models, DLBP1 and DLBP2, and compare them with other state-of-the-art deep learning-based prognostic models using a dataset from NASA data repository. All experiments are executed on a Dell R7425 server with an AMD Epyc 8 core @ 2.2 GHz and a 512GB RAM.

\subsection{Data Description}\label{section41}

The Commercial Modular Aero-Propulsion System Simulation (C-MAPSS) dataset has been widely used for the performance evaluation of prognostic models. The entire dataset consists of four sub-datasets, namely, FD001, FD002, FD003, and FD004. The difference among the four sub-datasets lies in the aircraft engines' flight conditions (i.e., throttle resolver angle, altitude, and ambient temperature) and the number of fault/failure modes (i.e., the fan and/or HPC degradation). In this article, we choose FD003 for model evaluation because it comprises data collected from engines with two failure modes (i.e., the failure of some engines is attributed to the fan degradation and others experience failure due to HPC degradation) and operates under a single flight condition (i.e., the flight condition remains constant throughout the duration of the degradation process). The dataset is composed of the following components: (i) degradation signals from 100 training engines that were run to failure, (ii) degradation signals from 100 test engines was prematurely terminated at random time points prior to their failure time, and (iii) the real RULs of the 100 test engines. We refer readers to \cite{saxena2008damage} for a more comprehensive introduction of the dataset.

\begin{table}[!t]\centering
\caption{Outline of the four sub-datasets. Conditions include the composition of a range of values for throttle resolver angle, altitude, and ambient temperature. Fault modes refer to fan degradation and HPC degradation.}
\vspace{2mm}
\begin{tabular}{lcccc}
\hline
 & \multicolumn{4}{c}{\text{C-MAPSS Dataset}} \\\cline{2-5} 
 &   FD001  &   FD002  &  FD003   & FD004    \\\hline
Training Engine Size&  100   &  260   &  100  & 248    \\
Testing Engine Size &  100   &   259  &  100   &   249  \\
Conditions &    1 & 6    & 1    &  6   \\
Fault Modes  &   1  & 1    &   2  &   2 \\
\hline
\label{table41}
\end{tabular}
\end{table}

\subsection{Data Preprocessing}\label{section42} 
The engines in FD003 are monitored by 21 sensors, which generate 21 time series-based degradation signals for each engine. Since the signals from sensors 1, 5, 16, 18, and 19 are constant throughout the engines' entire life cycle, they do not provide any useful health/degradation information and are thus excluded. As a result, there are 16 sensors (i.e., sensors 2, 3, 4, 6, 7, 8, 9, 10, 11, 12, 13, 14, 15, 17, 20, 21) left for the prognostic model construction. Using the notation in Section \ref{section32}, we have $P=16$ for $i=1,2,\cdots, N$ with $N=100$.

To ensure uniformity in the signal observations of each sensor within a consistent range, we use the min-max normalization, which scales the degradation signals from each sensor to $[0,1]$. Recall the degradation signals of asset $i$ is denoted by $\mathbf{X}_i\in\Real^{n_i\times P}$, where $n_i$ is the length of degradation signals of asset $i$, $P$ is the number of sensors, and $i=1,2,\ldots, N$. We let $\mathbf{X}_i = ({\boldsymbol{x}_i^{(1)}},{\boldsymbol{x}_i^{(2)}},\ldots,{\boldsymbol{x}_i^{(P)}})$, where $\boldsymbol{x}_i^{(j)}\in\Real^{n_i}$ is the degradation signal from the $j$th sensor of asset $i$, $j=1,2,\ldots,P$. As a result, the min-max normalization for sensor $j$ can be achieved by employing the following equation:

\begin{equation}\label{equation421}
\begin{aligned}
 \boldsymbol{x}_{\text{normal},i}^{(j)}=\frac{\boldsymbol{x}_i^{(j)}-\min{(\{\boldsymbol{x}_i^{(j)}\}_{i=1}^N)}}{\max((\{\boldsymbol{x}_i^{(j)}\}_{i=1}^N))-\min((\{\boldsymbol{x}_i^{(j)}\}_{i=1}^N)}.
    \end{aligned}
\end{equation}

The normalized signals from asset $i$ is $\mathbf{X}_{\text{normal},i} = ({\boldsymbol{x}_{\text{normal},i}^{(1)}},{\boldsymbol{x}_{\text{normal},i}^{(2)}},\ldots,{\boldsymbol{x}_{\text{normal},i}^{(P)}})$. As discussed in Section \ref{section321}, the SWM is applied to $\{\mathbf{X}_{\text{normal},i},y_i\}_{i=1}^N$ to generate degradation signals with the same length, and the generated degradation signals coupled with their corresponding RULs are $\{\tilde{\mathbf{X}}_{i},\tilde{y}_i\}_{i=1}^n$, where $n$ is the number of samples (``assets") after applying the SWM. Since degradation is observed only after the engine operates for a certain number of life cycles (there is no degradation during the early stages of asset life cycles), it is reasonable to maintain the RUL as a constant during the initial life cycles of a sample (representing a non-degradation status) \cite{heimes2008recurrent}. Following the suggestions of \cite{heimes2008recurrent}, \cite{ellefsen2019remaining}, and \cite{da2020remaining}, we set the number of constant life cycles during the early non-degradation stages as 125.

\subsection{Evaluation Metrics}\label{section43}

We use three criteria to evaluate the performance of our proposed method and to compare with benchmarks:
(i) Root Mean Square Error (RMSE), (ii) Prediction Score (PS), and (iii) Relative Absolute Error (RAE). The RMSE is calculated using Equation (\ref{equation431}), where $\delta_i$ is the difference between the predicted RUL (the output of the proposed DLBP models) and the real RUL of asset $i$, i.e., $\delta_i=f(\tilde{\mathbf{X}}_i;\vtheta_l,\vtheta_f,\vtheta_d)-\tilde{y}_i$, $i=1,\ldots,n_t$, where $n_t$ is the total number of samples in the test dataset (after applying the SWM). 

\begin{equation}\label{equation431}
\begin{aligned}
\text{RMSE} = \sqrt{\frac{1}{n_t}\sum_{i=1}^{n_t}\delta_i^2}
    \end{aligned}
\end{equation}

        \begin{figure}[!t]
         \centering
	\includegraphics[width=3.5in]{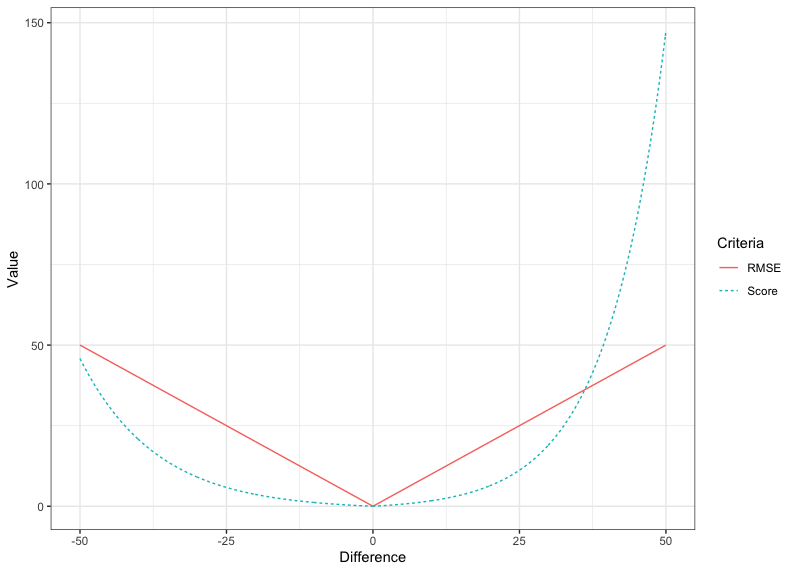}
	  \caption{A comparison of RMSE and PS with $x-$axis be the difference between the predicted response and the true value, and $y-$axis be the values of RMSE (in solid red) and PS (in dashed blue).
}
	\label{rmsescore}
     \end{figure}
     
Equation (\ref{equation432}) shows the calculation of the second evaluation metric, the PS, which is proposed by the article \cite{saxena2008damage} and has been used in many studies using the C-MAPSS dataset \cite{li2018remaining}\cite{kim2020bayesian}. A comparison of the RMSE and the PS is illustrated in Figure \ref{rmsescore}. The RMSE is symmetric, which gives an equal penalty to an underestimated RUL (the predicted RUL is smaller than the true RUL) and an overestimated RUL (the predicted RUL is larger than the true RUL). Such a symmetric penalty function exhibits limited efficacy in practical applications due to the increased severity of consequences stemming from an overestimated RUL, as an overestimated RUL indicates a higher chance of having an unexpected failure than an underestimated RUL. As indicated in Figure \ref{rmsescore}, the PS function is asymmetric and addresses this challenge by imposing a more substantial penalty for an overestimated RUL.
%is not effective in reality since usually the consequence of an overestimated RUL is more serious than an underestimated RUL. 
%This is because an overestimated RUL implies a higher probability of having an unexpected failure. As indicated in Figure \ref{rmsescore}, the PS function is asymmetric and addresses this challenge by penalizing more for an overestimated RUL.

\begin{equation}\label{equation432}
\begin{aligned}
\text{Score}_i =
    \begin{cases}
   e^{-\frac{\delta_i}{13}}-1 & \text{if $\delta_i<0$}\\
  e^{\frac{\delta_i}{10}}-1 & \text{if $\delta_i\geq 0$}
    \end{cases}       
    \end{aligned}
\end{equation}

The third evaluation metric is the RAE, which is computed using Equation (\ref{equation433}) below:

\begin{equation}\label{equation433}
\begin{aligned}
\text{RAE}_i = \frac{|\delta_i|}{|\tilde{y}_i|}
    \end{aligned}
\end{equation}

\subsection{DLBP Model Establishing, Training, and Evaluation}\label{section45}

\begin{table}[!t]
\centering
\caption{The tuning parameter candidates and their values.}
\begin{tabular}{ll}
\hline
Hyperparameter & Values \\ \hline
SWM &  \{15, 20, 25, 30, 35\}    \\
Number of LSTM Layers & \{1, 2\}      \\
Number of FC Layers &  \{1, 2\}  \\
LSTM Units &   \{64, 128, 256\}   \\
FC Units  &  \{32, 64, 128\}  \\
Batch Size & \{128, 256, 512\} \\
Number of Epochs & \{120, 150, 200, 250\}\\
\hline
\label{table42}
\end{tabular}
\end{table}

\begin{table*}[!t]
\scriptsize
\centering
\caption{The optimized tuning parameters and activation functions for model DLBP1 (AFDPL represents the Activation Functions in the Distribution Parameter Layer; \textcolor{black}{from left to right: location parameters; scale parameters; weights}).}
\begin{adjustbox}{minipage=18cm, center}
\begin{tabular}{lcccccc}
\hline
&  $T_w$  &  LSTM (units) & FC (units)   &  Batch Size & Epochs & AFDPL\\\hline
Mixture Log-Normal& 30& LSTM(128)  & FC(64)+FC(32) & 512  & 250  & \textit{elu}; \textit{softplus}; \textit{sigmoid}\\
Mixture Weibull &30 & LSTM(128)+LSTM(64) & FC(64) &  512 &250 & \textit{softplus}; \textit{softplus}; \textit{sigmoid}\\
Mixture Log-Logistic & 30&  LSTM(64)& FC(128) &  512 & 250  &\textit{softplus}; \textit{softplus}+1; \textit{sigmoid}\\
\hline
\label{table43}
\end{tabular}
\end{adjustbox}
\end{table*}

\begin{table*}[!t]
\scriptsize
\centering
\caption{The optimized tuning parameters and activation functions for model DLBP2 (AFDPL represents the Activation Functions in the Distribution Parameter Layer; \textcolor{black}{from left to right: location parameters; scale parameters; weights}).}
\begin{adjustbox}{minipage=18cm, center}
\begin{tabular}{lcccccc}
\hline
&  $T_w$  &  LSTM (units) & FC (units)   &  Batch Size & Epochs & AFDPL\\\hline
Mixture Log-Normal& 30& LSTM(256)+LSTM(128)  & FC(64) & 512  & 250  & \textit{elu}; \textit{$\cdot$}; \textit{sigmoid}\\
Mixture Weibull &25 & LSTM(256)+LSTM(64) & FC(128) &  512 &250 & \textit{softplus}; \textit{$\cdot$}; \textit{sigmoid}\\
Mixture Log-Logistic & 25&  LSTM(64)& FC(128)+FC(32) &  512 & 200  &\textit{softplus}; \textit{$\cdot$}; \textit{sigmoid}\\
\hline
\label{table45}
\end{tabular}
\end{adjustbox}
\end{table*}

We evaluate the performance of the proposed methods under three mixture LLS distributions: \textit{Mixture Log-Normal}, \textit{Mixture Weibull}, and \textit{Mixture Log-Logistic}. The FD003 dataset is used for training and testing, as discussed in Section \ref{section41}. For model training under each mixture distribution, stochastic gradient descent (SGD) is used to estimate the neuron weights and biases in the deep learning structure, while the Xavier initialization method \cite{glorot2010understanding} is applied for weight initialization. In addition to the weight and bias parameters, there are a certain number of tuning parameters that require optimization, including
%that need to be optimized. This includes 
(1) \textit{the window width (SWM step)}, (2) \textit{the number of LSTM layers}, (3) \textit{the number of FC layers}, (4) \textit{the number of units in each LSTM layer}, (5) \textit{the number of units in each FC layer}, (6) \textit{the batch size}, and (7) \textit{the number of epochs}. The batch size and the number of epochs are related to the SGD algorithm. The batch size controls the number of samples that are propagated through the network in each epoch (usually only a subset/batch of the training data is utilized by SGD in each epoch for the purpose of reducing the computation memory, accelerating the convergence speed, and avoiding over-fitting), and the number of epochs determines the number of iterations of the SGD algorithm. Table \ref{table42} summarizes the candidate values of each tuning parameter that have been investigated in this case study. 

Given the considerable number of tuning parameter combinations, estimating the parameters for each individual combination is computationally intensive. As an alternative, we employ a block-by-block optimization approach, where one block of tuning parameters is optimized at a time while keeping the other blocks fixed.
%it is computationally intensive to estimate the weight and bias parameters of the proposed model under each combination. Thus, we will optimize the tuning parameters block by block--that is--each time we will optimize one block tuning parameters while keeping other blocks fixed. 
To be specific, we first optimize \textit{the sliding window width} while setting \textit{the number of LSTM layers} as one, \textit{the number of FC layers} as two, \textit{the number of LSTM units} as 128, \textit{the number of neurons in the FC layer} as 64, \textit{the batch size} as 512, and \textit{the number of epochs} as 200. As illustrated in Table \ref{table42}, we have tried five widths for the sliding window: $15,20,25,30,35$. For each of the widths, we first apply the SWM on the training data in FD003 to generate training samples whose degradation signals have the same length. Then we randomly select $90\%$ of the training samples to estimate the weight and bias parameters and use the remaining $10\%$ data for model testing (RMSEs are computed). The training and testing process is repeated for $5$ times, and the average RMSE is calculated. The window width achieving the smallest average RMSE is chosen as the best one. Next, we optimize the second block of the tuning parameters: \textit{the number of LSTM layers} and \textit{the number of FC layers}. When optimizing the second block of hyperparameters, we set the first block (the window width) as the value that has been optimized earlier and set other blocks as the same values used when optimizing the first block. Since each of the two tuning parameters has two candidate values, there are in total four combinations if we consider them together: $\{1,1\},\{1,2\},\{2,1\},\{2,2\}$. For each combination, similarly, we split the training samples into training ($90\%$) and testing ($10\%$), repeat $5$ times, and use the average RMSE to select the best candidate. The third block of tuning parameters is the combinations of \textit{the number of units in each LSTM layer} and \textit{the number of units in each FC layer}, followed by the fourth block, which is \textit{the batch size} 
(see Table \ref{table42}). The last block to be optimized is \textit{the number of epochs}, which has four candidate values. We summarize the optimized tuning parameters for each of the three mixture distributions for the DLBP1 model in Table \ref{table43} and for the DLBP2 model in Table \ref{table45}. 

In addition to the tuning parameters, the activation function is another critical factor that significantly influences the performance of deep learning models. In this study, the state-of-the-art \textit{elu} \cite{clevert2015fast} activation function is used in the LSTM layers, the FC layers, and the distribution parameter layer. 
For the neurons with restricted output ranges, a suitable activation function should be selected accordingly. For example, for model DLBP1, if the mixture log-logistic distribution is used, the scale parameters have to be positive, as stipulated by the validity of the log-logistic distribution pdf. As a result, the output of the neurons corresponding to the scale parameters and the responses of the activation functions must also be positive.
In this article, the differentiable \textit{softplus} activation function is chosen for parameters whose range is $(0,\infty)$. The \textit{sigmoid} activation function is selected for the weights of the mixture distribution to guarantee the range is $(0,1)$. Tables \ref{table43} and \ref{table45} show the selected hyperparameters and activation functions of location parameters and scale parameters in this case study.

\subsection{Benchmarks, Results, and Analysis}\label{section46}
     \begin{figure}[!t]
         \centering
	\includegraphics[width=5.5in]{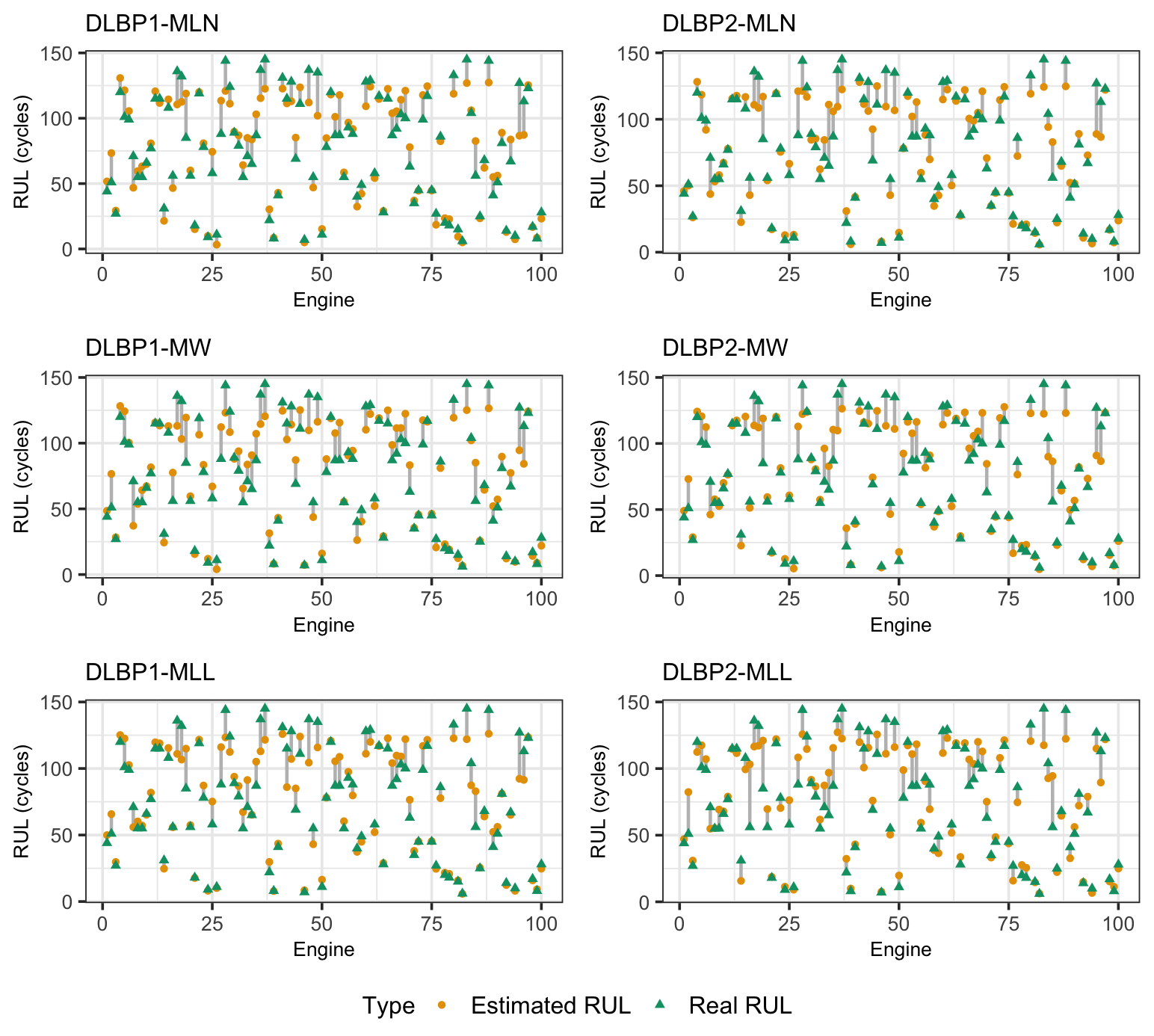}
\caption{\textcolor{black}{The estimated RUL values (orange circles) and actual RUL values (green triangles) for 100 test engines using the proposed methods are shown. The left three subplots represent Model DLBP1, and the right three subplots represent Model DLBP2, under three mixture distributions: mixture log-normal (top row), mixture Weibull (middle row), and mixture log-logistic (bottom row).}}
	\label{RULcomparison}
     \end{figure}

In this subsection, we demonstrate and analyze the performance of the proposed models and some state-of-the-art prognostic methods that use the same dataset for model validation. The results of the estimated RUL of 100 test engines compared with their real RUL values using the proposed methods are shown in Figure \ref{RULcomparison}.
As mentioned earlier, we evaluate the performance of our proposed model under three different mixture distributions: \textit{Mixture Log-Normal (MLN)}, \textit{Mixture Weibull (MW)}, and \textit{Mixture Log-Logistic (MLL)}. These distributions are denoted as “DLBP1-MLN,” “DLBP1-MW,” and “DLBP1-MLL” for DLBP1, and “DLBP2-MLN,” “DLBP2-MW,” and “DLBP2-MLL” for DLBP2.
We choose 11 existing models as baseline methods. The first baseline method is MODBNE proposed in the article \cite{zhang2016multiobjective}. This method is chosen because it is able to use multiple deep belief networks simultaneously, and, more importantly, its prediction accuracy outperforms many traditional data-driven methods such as SVM,
%\cite{smola2004tutorial}
LASSO,
gradient boosting,
and random forest.
\textcolor{black}{
Benchmark 2, referred to as “LSTM+NN”,  was proposed in \cite{zheng2017long}. It features a framework comprising a sequence of LSTM layers, with their output fed into feedforward neural networks. 
Benchmark 3 denoted as “Deep CNN” \cite{li2018remaining}, is a deep CNN designed for RUL prediction.
Benchmark 4, designated as “Semi-supervised DL” \cite{ellefsen2019remaining}, includes an unsupervised pre-training step followed by multiple LSTM layers. It achieves better performance compared to Benchmark 2.
Baseline method 5, “Bayesian LSTM”, is a Bayesian deep learning framework validated to outperform many existing deep learning-based prognostic models \cite{kim2020bayesian}.
Benchmark 6, referred to as “MS-DCNN” \cite{li2020remaining}, employs an architecture designed to extract multi-scale information from input degradation signals.
Baseline method 7, “ATS2S” \cite{ragab2021attention}, introduces an attention-based encoder-decoder framework that integrates features from both the encoder and decoder to predict RUL.
Benchmark 8 \cite{liu2022aircraft} utilizes a double attention-based network comprising a channel attention-based CNN and a Transformer to extract features and perform RUL prognostics.
Baseline method 9, “DS-SANN” \cite{xu2022novel}, proposes a dual-stream structure consisting of original data and transformed auxiliary data, combined with a multi-head self-attention mechanism for RUL estimation.
Benchmark model 10 ``HMCL" \cite{sharma2023hybrid} introduces a framework that combines multi-scale CNN and LSTM, achieving superior RMSE performance.
Benchmark 11, ``NBLSTM" \cite{al2020nblstm}, integrates a noisy CNN and a noisy bidirectional BLSTM in parallel to extract spatial and temporal features from the data for RUL prediction.}
Similar to our proposed model, the SWM is first applied to all the benchmark models as well. 
Despite the sliding window width varies across algorithms, the selection of a best-fit value ensures the performance of each model remains comparable.

\begin{table}[!t]
\color{black}
\centering
\caption{RMSE and PS of the proposed methods and the state-of-the-art benchmarks.}
\begin{tabular}{llcc}
\hline
No.& Algorithm[Ref.] &RMSE & Score \\ \hline
1&   MODBNE  \cite{zhang2016multiobjective}    & 12.51 & 521.91\\
  2&  LSTM+NN  \cite{zheng2017long} & 16.18 & 852 \\
3& Deep CNN  \cite{li2018remaining}  & 12.64 & 284.1 \\
4& Semi-supervised DL  \cite{ellefsen2019remaining} & 12.10& 251 \\
 5&Bayesian LSTM  \cite{kim2020bayesian}& 12.07 &409.39\\
  6&MS-DCNN  \cite{li2020remaining} & {11.67}& 241.89 \\
7&ATS2S  \cite{ragab2021attention} & \textbf{11.44} &263\\
8& Double Attention \cite{liu2022aircraft}
&13.39&290\\ 
9&DS-SANN  \cite{xu2022novel}&12.28 &286.07\\
10&HMCL \cite{sharma2023hybrid}&\textbf{9.75} & 280.62\\
11&NBLSTM\cite{al2020nblstm}&\textbf{11.36}&226.48\\
  
12&Proposed DLBP1-MLN & 11.97 & \textbf{208.25} \\
13&Proposed DLBP1-MW  & 12.7 & 220.23 \\
14&Proposed DLBP1-MLL  & \textbf{11.47} & \textbf{180.63}\\
15&Proposed DLBP2-MLN & 12.34 &  228.65\\
16&Proposed DLBP2-MW  & 11.84 & \textbf{187.64}\\
17&Proposed DLBP2-MLL  & 13.18 & 225.07\\
\hline
\label{table44}
\end{tabular}
\color{black}
\end{table}

\color{black}
We report the RMSE and PS values for our proposed methods and the 11 benchmark models in Table \ref{table44}, with the three smallest RMSE and PS values highlighted in bold. Table \ref{table44} demonstrates that our proposed models outperform all the benchmarks with respect to PS. Notably, the lowest three PS values are achieved by the proposed models “DLBP1-MLL”, “DLBP2-MW”, and “DLBP1-MLN”, with PS values of $180.63$, $187.64$, and $208.25$, respectively. Furthermore, our methods also secure the 4th, 5th, and 7th lowest PS values, recorded at $220.23$, $225.07$, and $228.65$, respectively. Among the benchmarking models, the lowest PS value is recorded by the NBLSTM model \cite{al2020nblstm}, with a value of $226.48$.
As previously discussed, PS is a superior metric for evaluating the performance of a prognostic model compared to RMSE, as it places greater penalties on overestimated RUL values than underestimated ones. This distinction is critical because overestimated RUL (i.e., predicted RUL is greater than the true RUL) carries a higher risk of unexpected failures, potentially leading to severe consequences.
In terms of RMSE, Table \ref{table44} shows that the HMCL benchmark achieves the lowest RMSE. However, its PS value is significantly higher than those of our proposed models. Notably, our DLBP1-MLL model achieves a RMSE of $11.47$, which can be seen as the second place (comparable to the ATS2S \cite{ragab2021attention} and NBLSTM \cite{al2020nblstm}).
The superior performance of our proposed models can be attributed to the incorporation of domain knowledge, specifically the mixture LLS distribution, into the deep learning framework. Unlike our approach, the benchmark models rely solely on deep learning techniques without leveraging domain knowledge in reliability engineering or survival analysis. 
\color{black}

   \begin{figure}[!t]
         \centering
	\includegraphics[width=5.5in]{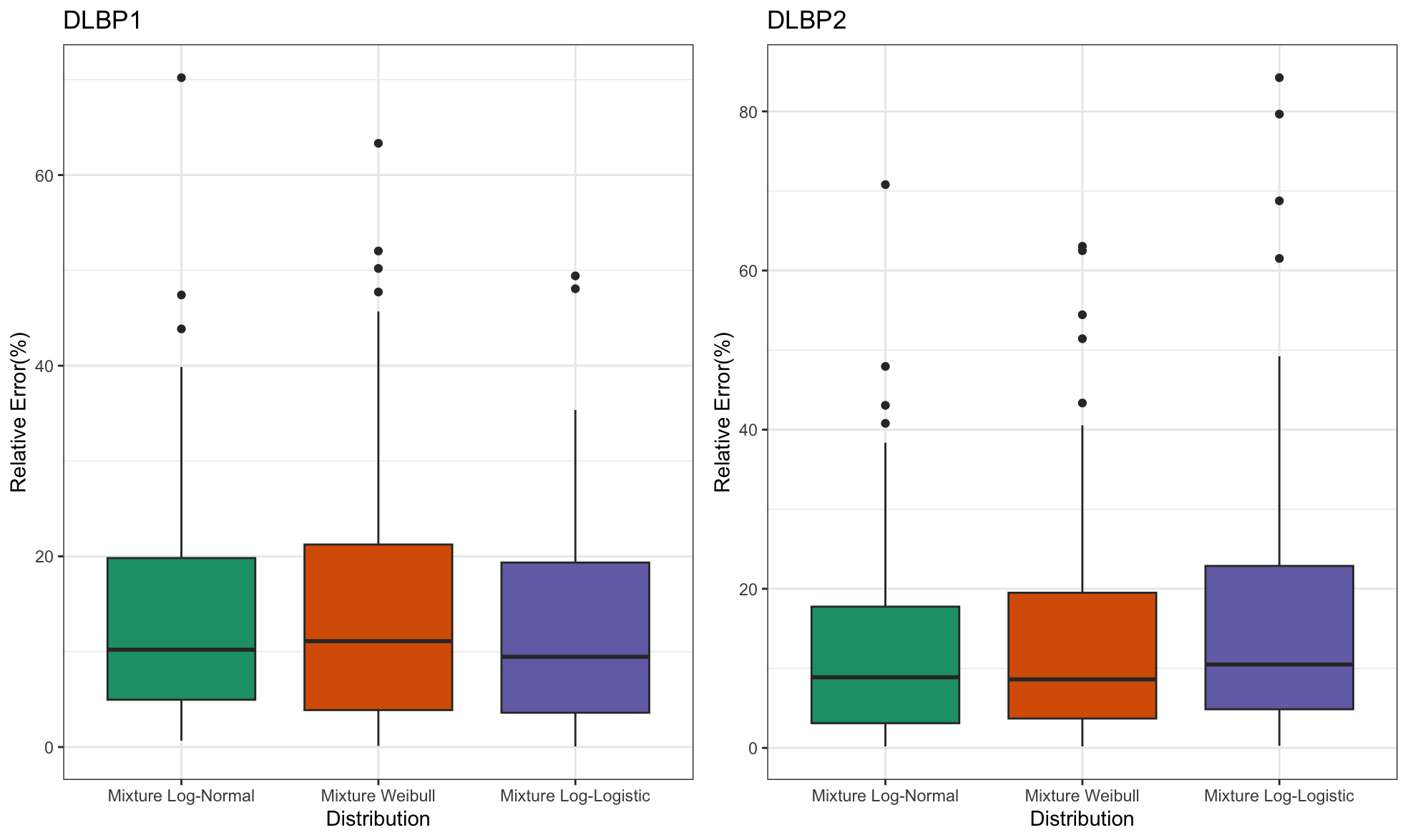}
	  \caption{The boxplot of RAE for the proposed methods (left: model DLBP1; right: model DLBP2) under three mixture distributions (from left to right in each subplot: Mixture Log-Normal, Mixture Weibull, Mixture Log-Logistic).}
	\label{REpercentage}
     \end{figure}

To further demonstrate the prediction performance of the proposed model, we report the RAE of DLBP1 and DLBP2 under the three distributions in Figure \ref{REpercentage}. Similar to RMSE, RAE treats underestimation and overestimation equally, suggesting its limited effectiveness compared with the PS metric. Nonetheless, RAE still provides some useful insights in evaluating the performance of the proposed model, and it is a widely used metric in performance evaluation of prognostic models that has been adopted by many other research works \cite{liu2013data,fang2017multistream,fang2021multi}. Figure \ref{REpercentage} shows that the ``DLBP2-MLL" model outperforms the other five models in terms of both prediction accuracy and precision, as evidenced by the medians and the interquartile range (IQRs) of the boxplots. In practice, while failure time often adheres to a distribution within the LLS family, the specific distribution to use is usually uncertain. This issue can be resolved through the use of model selection criteria. For instance, one might initially divide the historical data into training and validation datasets. Subsequently, various models are fitted on the training dataset and evaluated using the validation dataset. The model that yields the smallest validation error is then selected as the most suitable model.

\color{black}
\section{Conclusions}\label{section5}
\textcolor{black}{
In this article, we proposed two deep learning-based prognostic models for the RUL prediction of complex engineering assets with uncertain failure modes. Both models incorporate a mixture LLS distribution into a deep learning framework. The first model (i.e., DLBP1) allows each asset to have its own location parameters and scale parameters, while the second model (i.e., DLBP2) assumes shared scale parameters across all assets but allows for different location parameters, which is a common assumption in statistical learning. 
A dataset from the degradation of aircraft engines from the NASA data repository was used to validate the effectiveness of these two proposed models and evaluate their performance. Eleven state-of-the-art deep learning-driven prognostic methods were chosen as benchmarks. Numerical results demonstrate that the proposed deep learning-based prognostic models outperform the benchmarks in RUL prediction. The results demonstrate that the proposed models achieve the lowest PS among the benchmarks, a metric that better reflects the practical consequences of overestimated RUL values due to its asymmetric penalization. This highlights the effectiveness of the proposed models in applications where overestimation risks can lead to severe failures.
Furthermore, the integration of the proposed mixture LLS distribution with a deep learning framework addresses three critical challenges faced by existing prognostic models: overlapping signals due to ambiguous degradation sources, lack of labeled training data, and the similarity of degradation signals among components or failure modes. These strengths demonstrate the capability and adaptability of the proposed models, making them a valuable contribution to the field of prognostics and health management for complex systems. One potential limitation of the proposed methods is the limited flexibility of LSTMs in capturing complex correlation structures within the data. Expanding the proposed approach to incorporate more advanced deep learning techniques, such as those based on self-attention mechanisms, could be an interesting direction for future research.}

\color{black}

 \bibliographystyle{elsarticle-num} 
 \bibliography{cas-refs}





% \end{thebibliography}
\end{document}